\documentclass[11pt,a4paper]{article}
\usepackage[hyperref]{acl2019}
\usepackage{times}
\usepackage{latexsym}
\usepackage{tikz}
\usepackage{url}
\usepackage{multirow}
\usepackage[utf8]{inputenc}
\usepackage{enumitem}
\definecolor{mypink}{RGB}{255, 160, 160}
\hypersetup{colorlinks=true, urlcolor=blue}

\usepackage[colorinlistoftodos]{todonotes} 

\aclfinalcopy 

\title{Improving Chemical Named Entity Recognition in Patents \\ with Contextualized Word Embeddings}

\author{ Zenan Zhai$^1$, Dat Quoc Nguyen$^1$, Saber A. Akhondi$^2$, Camilo Thorne$^2$, \\ \textbf{Christian Druckenbrodt$^2$, Trevor Cohn$^1$, Michelle Gregory$^2$, Karin Verspoor$^1$}\\
$^1$The University of Melbourne, Australia; $^2$Elsevier\\
${}^{1}${\small \tt{\{zenan.zhai,dqnguyen,trevor.cohn,karin.verspoor\}@unimelb.edu.au}}\\
${}^{2}${\small \tt{\{s.akhondi,c.thorne.1,c.druckenbrodt,m.gregory\}@elsevier.com}}
}

\date{}

\begin{document}
\maketitle
\begin{abstract}
Chemical patents are an important resource for chemical information.
However, few chemical Named Entity Recognition (NER) systems have been evaluated on patent documents, due in part to their structural and linguistic complexity. 
In this paper, we explore the NER performance of a BiLSTM-CRF model utilising pre-trained word embeddings, character-level word representations and contextualized ELMo word representations for chemical patents. We compare word embeddings pre-trained on biomedical and chemical patent corpora. The effect of tokenizers optimized for the chemical domain on NER performance in chemical patents is also explored. The results on two patent corpora show that contextualized word representations generated from ELMo substantially improve chemical NER performance w.r.t.\ the current state-of-the-art. 
We also show that domain-specific resources such as word embeddings trained on chemical patents and chemical-specific tokenizers have a positive impact on NER performance.
\end{abstract}


\section{Introduction}

Chemical patents are an important starting point for understanding of chemical compound purpose, properties, and novelty.
New chemical compounds are often initially disclosed in patent documents; however it may take 1-3 years for these chemicals to be mentioned in chemical literature \citep{senger2015managing}, suggesting that patents are a valuable but underutilized resource. 
As the number of new chemical patent applications is drastically increasing every year \citep{muresan2011making}, it is becoming increasingly important to develop automatic natural language processing (NLP) approaches enabling information extraction from these patents \citep{akhondi2019automatic}. 
Chemical Named-Entity Recognition (NER) is a fundamental step for information extraction from chemical-related texts, supporting relation extraction \citep{wei2016assessing}, reaction prediction \citep{schwaller2018found} and retro-synthesis \citep{segler2018planning}. 

However, performing NER in chemical patents can be challenging \citep{akhondi2014annotated}. As legal documents, patents are written in a very different way compared to scientific literature. When writing scientific papers, authors strive to make their words as clear and straight-forward as possible, whereas patent authors often seek to protect their knowledge from being fully disclosed \citep{valentinuzzi2017patents}. 

In tension with this is the need to claim broad scope for intellectual property reasons, and hence 
patents typically contain more details and are more exhaustive than scientific papers \citep{lupu2011current}. 

There are a number of characteristics of patent texts that create challenges for NLP in this context. Long sentences listing names of compounds in chemical patents are frequently used. The structure of sentences in patent claims is usually complex, and syntactic parsing in patents can be difficult \cite{hu2016improving}.  A quantitative analysis by \citet{verberne2010quantifying} showed that the  average sentence length in a patent corpus is much longer than in general language use. That work also showed that the lexicon used in patents usually includes domain-specific and novel terms that are difficult to understand.
Some patent authorities use Optical Character Recognition (OCR) for digitizing patents, which can be problematic when applying automatic NLP approaches as the OCR errors introduces extra noise to the data \cite{akhondi2019automatic}.

Most NER systems for the chemical domain were developed, trained and tested on either chemical literature or only the title and abstract of chemical patents \citep{akhondi2019automatic}. There are substantial linguistic differences between abstracts and the corresponding full text publications \citep{Cohen2010}. The performance of NER approaches on full patent documents has still not been fully explored \citep{krallinger2015overview}. 

Hence, this paper will focus on presenting the best NER performance achieved to date on full chemical patent corpus. 

We use a combination of pre-trained word embeddings, a CNN-based character-level word representation and contextualized word representations generated from ELMo, trained on a patent corpus, as input to a BiLSTM-CRF model. The results show that contextualized word representations help improve chemical NER performance substantially.
In addition, the impact of the choice of pre-trained word embeddings and tokenizers is assessed.

The results show that  word embeddings that are pre-trained on chemical patents outperform embeddings pre-trained on biomedical datasets, and using tokenizers optimized for the chemical domain can improve NER performance in chemical patent corpora.

\section{Related work}

In this section, we summarize previous methods and empirical studies on NER in chemical patents.

Two existing Conditional Random Field (CRF)-based systems for chemical named entity recognition are tmChem \citep{Leaman:2015}  and ChemSpot \citep{rocktaschel2012chemspot}; each makes use of numerous hand-crafted features including word shape, prefix, suffix, part-of-speech and character N-grams in an algorithm based on modelling of tag sequences. 
A previous detailed empirical study
 explored the generalization performance of these systems and their ensembles \cite{habibi2016recognizing}. The application of the tmChem model trained on chemical \textit{literature} corpora of the BioCreative IV CHEMDNER task \cite{krallinger2015overview} and the ChemSpot model trained on a subset of the SCAI corpus \citep{klinger2008detection} 
 resulted in a significant performance drop over chemical \textit{patent} corpora.

\citet{zhang2016chemical} compared the performance of CRF- and Support Vector Machine (SVM)-based models on the CHEMDNER-patents corpus \cite{krallinger2015overview}. The features constructed in that work included the binarized embedding \citep{guo2014revisiting}, Brown clustering \citep{brown1992class} and domain-specific features extracted by detecting common prefixes/suffixes in chemical words. The obtained results show that the performance of CRF and SVM models can be significantly improved by incorporating unsupervised features (e.g.\ word embeddings, word clustering). The study also showed that the SVM model slightly outperformed the CRF model in the chemical NER task.

To perform chemical NER on the CHEMDNER~patents corpus, \citet{akhondi2016chemical} proposed an ensemble approach combining  a gazetteer-based method and a modified version of tmChem. Here, the  gazetteer-based method utilized a wide range of chemical dictionaries, while  additional features such as stems, prefixes/suffixes, chemical elements were added to the original feature set of tmChem. In the ensemble approach, tokens were predicted as chemical mentions if recognized as positive by either tmChem or the gazetteer-based method. The results showed that both gazetteer-based and ensemble approaches were outperformed by the modified tmChem version in terms of overall $F_1$ score, although these two approaches can obtain higher recall.

\citet{huang2015bidirectional} proposed a BiLSTM-CRF based on the use of a bidirectional long-short term memory network -- BiLSTM  \citep{schuster1997bidirectional} -- to extract (latent) features for a CRF classifier. The BiLSTM encodes the input in both forward and backward directions and passes the concatenation of outputs from both directions as input to a linear-chain CRF sequence tagging layer. In this approach, the BiLSTM selectively encodes information and long-distance dependencies observed while processing input sentences in both directions, while the CRF layer globally optimizes the model by using information from neighbor labels. 

The morphological structures within words are also important clues for identifying named entities in biological domain. Such morphological structures are widely used in systematic chemical name formats (e.g.\ IUPAC names) and hence particularly informative for chemical NER \citep{klinger2008detection}. Character-level word representations have been developed to leverage information from these structures by encoding the character sequences within tokens. \citet{Ma2016endtoend} uses Convolutional Neural Networks (CNNs) to encode character sequences 
while \citet{Lample:2016} developed a LSTM-based approach for encoding character level information.

\citet{Maryam2017deep} presented an empirical study comparing three NER models on a large collection of biomedical corpora including the BioSemantics patent corpus: (1) tmChem--the CRF-based model with hand-crafted features--used as the baseline; (2) a second CRF model based on CRFSuite \citep{okazaki2007crfsuite} using pre-trained word embeddings; (3) and a BiLSTM-CRF model with additional LSTM-based character-level word embeddings \citep{Lample:2016}. The performance of CRFSuite- and BiLSTM-CRF-based models with different sets of  pre-trained biomedical word embeddings \citep{moen2013distributional} were also explored. The results showed that the BiLSTM-CRF model with the combination of domain-specific pre-trained word embedding and LSTM-based character-level word embeddings outperformed the two CRF-based models on chemical NER tasks in both chemical literature and chemical patent corpora. However, this work used only a general tokenizer (i.e. OpenNLP) and word embeddings pre-trained on biomedical corpora.

\citet{corbett2018chemlistem} presented word-level and character-level BiLSTM networks for chemical NER in literature domain. The word-level model employed  word embeddings learned by GloVe~\citep{pennington2014glove} on a corpus of patent titles and abstracts. The character-level model used two different transfer learning approaches to pre-train its character-level encoder. The first approach attempts to predict neighbor characters at each time step, while the other tries to predict whether a given character sequence is an entry in the chemical database ChEBI~\citep{degtyarenko2007chebi}. Experimental results show that the character-level model can produce better NER performance than word-level model by leveraging transfer learning. In addition, for the word-level  model, using pre-trained word embeddings learned from a patent corpus helps produce better performance than using the pre-trained ones learned from a general corpus.

\section{Our empirical methodology}

This section presents our empirical study of  NER chemical on patent datasets. We first outline the experimental datasets (Section \ref{ssec:data}) and the tokenizers (Section \ref{ssec:tokenizer}) used to pre-process these datasets, and then we  introduce the  BiLSTM-CRF-based models (Section \ref{ssec:models}) with pre-trained word embeddings (Section \ref{ssec:word}),  character-level word embeddings (Section \ref{ssec:chars}),  contextualized word embeddings (Section \ref{ssec:elmo}) and implementation details (Section \ref{ssec:implement}). 

\subsection{Dataset}\label{ssec:data}

We conduct experiments on 2 patent corpora: the BioSemantics patent corpus \citep{akhondi2014annotated} and Reaxys gold set \citep{akhondi2019automatic}.

The BioSemantics patent corpus \citep{akhondi2014annotated} consists of 200 full chemical patent documents with 9 different entity classes. In particular, this corpus has 170K sentences and and 360K entity annotations, which is much larger than previously used datasets, e.g.\ the CHEMDNER patent abstract corpus \cite{krallinger2015overview}.
Therefore, this corpus can be considered as a more suitable resource for evaluating deep learning methods in which a large amount of training data is required \citep{lecun2015deep}. A subset of 47 patents were annotated by multiple groups (at least 3) of annotators and evaluated through inner-annotator agreement. 
By harmonizing the annotations from different annotator groups, these 47 patents formed the ``\textit{harmonized}'' set in the BioSemantics patent corpus. We use the harmonized set for both hyper-parameter tuning and error analysis as it has known high-quality annotations.

The Reaxys gold set \citep{akhondi2019automatic} contains 131 patent snippets (parts of full chemical patent documents) from several different patent offices. The tagging scheme of this corpus includes 2 coarse-grained labels chemical class and chemical compounds, and 7 fine-grained labels of chemical compound (e.g.\ \textit{mixture-part}, \textit{prophetic}) and chemical class (e.g.\ \textit{bio-molecule}, \textit{
Markush}, \textit{mixture}, \textit{mixture-part}). This corpus is relatively small in size, approximately 20,000 sentences in total, but very richly annotated. The relevancy score of each chemical entity and the relations between them were also annotated, which allows this corpus to be used in other tasks beyond named entity recognition.

In our experiments, each corpus is used separately. We follow \citet{Maryam2017deep} to use a ratio split of 60\%/10\%/30\% for training/development/test. Note that on the BioSemantic patent corpus,  our sampling of datasets may not be exactly the same as in \citet{Maryam2017deep}.

\subsection{Tokenizers}\label{ssec:tokenizer}

The morphological information captured by character-level word representations can be highly affected by tokenization quality. General-purpose tokenizers usually split tokens by spaces and punctuation. However, strict adherence to such boundaries 
may not be suitable for chemical texts as spaces and punctuation are commonly used in the IUPAC format for chemical names (e.g.\ \textit{3-(4,5-dimethylthiazol-2-yl)-2,5-diphenyl tetrazolium bromide}) \citep{jessop2011oscar4}. Hence, the impact of different tokenizers on NER also needs to be explored.

A pre-processing step is applied to the patent corpora including sentence detection and tokenization. Following \citet{Maryam2017deep}, we use the OpenNLP \citep{morton2005opennlp} English sentence detection model. To explore the relationship between tokenization quality and final NER performance, we apply different tokenizers and train/test models with each tokenizer individually.  To investigate the effect of a general domain tokenizer, following \citet{Maryam2017deep}, we also use the OpenNLP tokenizer. To investigate whether NER performance will be affected by tokenization quality, we employ three tokenizers optimized for chemical texts including ChemTok \citep{akkasi2016chemtok}, OSCAR4  \citep{jessop2011oscar4} and NBIC UMLSGeneChemTokenizer.\footnote{NBIC UMLSGeneChemTokenizer is  developed by the Netherlands Bioinformatics Center, available at  \url{https://trac.nbic.nl/data-mining/wiki}.}

\begin{figure}[!t]
    \centering
    \resizebox{.5\textwidth}{!}{
    \newcommand{\trinode}[6] {%
    \node (#4) at (0,0) [draw, align=center] {#1};
    \node (#5) at (-1.5,0) [draw, align=center] {#2};
    \node (#6) at (1.5,0) [draw, align=center] {#3};
}

\begin{tikzpicture}
\tikzset{
 compnode/.style={draw,minimum size=1em,line width=1pt,circle}
}
\pgfsetarrowsend{latex}
\matrix[column sep=4ex, row sep=2ex]{
   & \node (tag0) {B-PER};
   & \node (tag1) {O};
   & \node (tag2) {B-LOC};
   \\
   & \node[draw, minimum width=1.5cm] (crf0) {CRF};
   & \node[draw, minimum width=1.5cm] (crf1) {CRF};
   & \node[draw, minimum width=1.5cm] (crf2) {CRF};
   \\
   & \node[draw, align=center, minimum width=1.5cm] (l0) {Linear \\ Layer};
   & \node[draw, align=center, minimum width=1.5cm] (l1) {Linear \\ Layer};
   & \node[draw, align=center, minimum width=1.5cm] (l2) {Linear \\ Layer};
   \\
   & \node (c10) {\Huge $\oplus$};
   & \node (c11) {\Huge $\oplus$};
   & \node (c12) {\Huge $\oplus$};
   \\
   & \node[draw, minimum width=1.5cm] (h20) {$LSTM$};
   & \node[draw, minimum width=1.5cm] (h21) {$LSTM$};
   & \node[draw, minimum width=1.5cm] (h22) {$LSTM$};
   \\
   \\
   & \node[draw, minimum width=1.5cm] (h10) {$LSTM$};
   & \node[draw, minimum width=1.5cm] (h11) {$LSTM$};
   & \node[draw, minimum width=1.5cm] (h12) {$LSTM$};
   \\
   & \node (c00) {\Huge $\oplus$};
   & \node (c01) {\Huge $\oplus$};
   & \node (c02) {\Huge $\oplus$};
   \\
   & \trinode{Word \\ emb.}{CNN \\ char.}{ELMo \\ emb.}{x00}{x01}{x02};
   & \trinode{Word \\ emb.}{CNN \\ char.}{ELMo \\ emb.}{x10}{x11}{x12};
   & \trinode{Word \\ emb.}{CNN \\ char.}{ELMo \\ emb.}{x20}{x21}{x22};
   \\
   & \node (i0) {Mark};
   & \node (i1) {visited};
   & \node (i2) {Paris};
   \\
 };

 \path
 (i0) edge (x00)
 (i0) edge (x01)
 (i0) edge (x02)
 (c00) edge (h10)
 (h20) edge (c10)
 (c10) edge (l0)
 (l0) edge (crf0)
 (crf0) edge (tag0)
 (i1) edge (x10)
 (i1) edge (x11)
 (i1) edge (x12)
 (c01) edge (h11)
 (h21) edge (c11)
 (c11) edge (l1)
 (l1) edge (crf1)
 (crf1) edge (tag1)
 (i2) edge (x20)
 (i2) edge (x21)
 (i2) edge (x22)
 (c02) edge (h12)
 (h22) edge (c12)
 (c12) edge (l2)
 (l2) edge (crf2)
 (crf2) edge (tag2)
 (h11) edge (h12)
 (h10) edge (h11)
 (h22) edge (h21)
 (h21) edge (h20)
;

\path[-]
 (crf0) edge (crf1)
 (crf1) edge (crf2)
;

\draw (x01) |- (c00);
\draw (x00) -- (c00);
\draw (x02) |- (c00);

\draw (x11) |- (c01);
\draw (x10) -- (c01);
\draw (x12) |- (c01);

\draw (x21) |- (c02);
\draw (x20) -- (c02);
\draw (x22) |- (c02);

 \path[bend left=60]
 (h10) edge (c10)
 (c00) edge (h20)
 (h11) edge (c11)
 (c01) edge (h21)
 (h12) edge (c12)
 (c02) edge (h22)
;

\end{tikzpicture}
    }
    \caption{Architecture of EBC-CRF}
    \label{fig:model}
\end{figure}
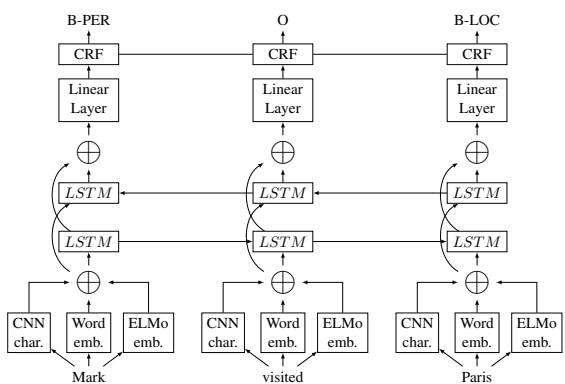

\subsection{Models}\label{ssec:models}

We use  the BiLSTM-CNN-CRF model \citep{Ma2016endtoend}  as our baseline.  We extend the baseline by adding the contextualized word representations generated from ELMo \citep{peters2018deep}. For convenience, we call the extended version as EBC-CRF as illustrated in Figure \ref{fig:model}.
In particular, for EBC-CRF, we use a concatenation of pre-trained word embeddings, CNN-based character-level word embeddings and ELMo-based contextualized word embeddings as the input of a BiLSTM encoder. The BiLSTM encoder learns a latent feature vector for each word in the input. Then each latent feature vector is linearly transformed before being fed into a linear-chain CRF layer \citep{Lafferty:2001} for NER tag prediction. 
We assume binary potential between tags and unary potential between tags and words.

\subsection{Pre-trained word embeddings}\label{ssec:word}

\citet{dai2019} showed that NER performance is significantly affected by the overlap between pre-trained word embedding vocabulary and the target NER data. Therefore, we explore the effects of different sets of pre-trained word embeddings on the NER performance.

We use 200-dimensional pre-trained PubMed-PMC and Wiki-PubMed-PMC word embeddings \citep{moen2013distributional}, which are widely used for NLP tasks in biomedical domain.
Both the PubMed-PMC and Wiki-PubMed-PMC embeddings word embeddings were generated by training the Word2Vec skip-gram model \citep{mikolov2013skipgram} on a collection of PubMed abstracts and PubMed Central articles. Here, an additional Wikipedia dump was also used to learn the Wiki-PubMed-PMC word embeddings. 

To explore whether word embeddings trained in the same domain can produce better performance in NER tasks, we learn another set of word embeddings, which we called \textit{ChemPatent} embeddings, by applying the same model and hyper-parameters from \citet{moen2013distributional} on a collection of 84,076 full patent documents (1B tokens) across 7 patent offices (see Table \ref{tab:elmo_statistics} for details).

The pre-trained PubMed-PMC, Wiki-PubMed-PMC and ChemPatent word embeddings are fixed during training of the NER models. For a more concrete comparison, a set of 200-dimensional trainable word embeddings initialized from normal distribution is used as a baseline. 

The 200-dimensional baseline word embeddings contain  all words in the vocabulary of the dataset and are initialized from a normal distribution, the baseline word embeddings are learned during training process. The vocabulary of models using pre-trained word embeddings is  built by taking the union of words in the pre-traied word embedding file and words with frequency more than 3 in training and development sets. We do not update weights for word embeddings if pre-trained word embeddings were used.

\subsection{Character-level representation}\label{ssec:chars}

The BiLSTM-CRF model with character-level word representations \citep{Lample:2016, Ma2016endtoend} has been shown to have state-of-the-art performance in NER tasks on chemical patent datasets \citep{Maryam2017deep}. It has been shown that the choice of using LSTM-based or CNN-based character-level word representation has little effect on final NER performance in both general and biomedical domain while the CNN-based approach has the advantage of reduced training time \citep{Nils2017reporting, zhai2018comparing}.
Hence, we use the CNN-based approach with the same hyper-parameter settings of \citet{Nils2017reporting} for capturing character-level information (see Table~\ref{tab:hyper-parameters} for details).

\subsection{ELMo} \label{ssec:elmo}

\begin{table}[!t]
    \centering
    \scalebox{0.8}{
    \begin{tabular}{l|l|l|l}
    \hline
    \bf Patent Office & \bf Document & \bf Sentence & \bf Tokens  \\
    \hline
    AU & 7,743 & 4,662,375 & 156,137,670\\
    CA & 1,962 & 463,123 & 16,109,776 \\
    EP & 19,274 & 3,478,258 & 117,992,191\\
    GB & 918 & 182,627 & 6,038,837\\
    IN & 1,913 & 261,260 & 9,015,238\\
    US & 41,131 & 19,800,123 & 628,256,609\\
    WO & 11,135 & 4,830,708 & 159,286,325\\
    \hline
    Total & 84,076 & 33,687,474 & 1,092,836,646 \\
    \hline
    \end{tabular}
    }
    \caption{Statistics of the unannotated patent corpus used for training ChemPatent embeddings and ELMo.}
    \label{tab:elmo_statistics}
\end{table}

ELMo \citep{peters2018deep} and BERT \citep{devlin2018bert} can be used to generate contextualized word representations by combining internal states of different layers in neural language models. Contextualized word representation can help to improve performance in various NLP tasks by incorporating contextual information, 
essentially allowing for the same word to have distinct context-dependent meanings. This could be particularly powerful for chemical NER since generic chemical names (e.g.\ \textit{salts, acid}) may have different meanings in other domains. We therefore explore the impact of using contextualized word representations for chemical patents. 

We train ELMo on the same corpus of 84K patents  (detailed in Table \ref{tab:elmo_statistics}), which we use for training the ChemPatent embeddings  (described in Section \ref{ssec:word}). 
We use the ELMo implementation provided by~\citet{peters2018deep} with default hyper-parameters.\footnote{\url{https://github.com/allenai/bilm-tf}} 
Such neural language models require a large amount of computational resources to train. 
In ELMo, a maximum character sequence length of tokens is set to make training feasible. However, systematic chemical names in chemical patents are often longer than the typical maximum sequence length of these neural language models. 
As very long tokens tend to be systematic chemical names, we reduced the max length of word from 50 to 25 and replace tokens longer than 25 characters by a special token ``Long\_Token''.

\subsection{Implementation details}\label{ssec:implement}

\begin{table}[!t]
    \centering
        \scalebox{0.75}{
        \begin{tabular}{cc}
         \begin{tabular}{|l|l|}
        \hline \bf Hyper-para. & \bf Value \\ \hline
         Optimizer & Adam \\
         Learning rate & 0.001 \\
         Mini-batch size & 16 \\
         Clip Norm(L2) & $1$  \\
         Dropout & [0.25, 0.25] \\
         \hline
        \end{tabular}
             & 
        \begin{tabular}{|l|l|}
        \hline
        \textbf{Hyper-para.} & \textbf{Value}\\\hline
        charEmbedSize & 50 \\
        filter length & 3 \\
        \# of filters & 30 \\
        \hline
        output size & 30 \\
        \hline
        \end{tabular}
        \\
        \\
        (a) BiLSTM-CRF &  (b) CNN-char \\
        \end{tabular}
        }
    \caption{Fixed hyper-parameter configurations.}
    \label{tab:hyper-parameters}
\end{table}

Our NER model implementation is based on the AllenNLP system~\citep{Gardner2017AllenNLP}. We learn model parameters using  the training set, and we use the overall $F_1$ score over development set as indicator for performance improvement. All models in this paper are trained with 50 epochs in maximum, and an early stopping is  applied if there are no overall $F_1$ score improvement observed after 10 epochs.   

In~\citet{Nils2017optimal} and~\citet{zhai2018comparing}, optimal hyper-parameters of BiLSTM-CRF models in NER tasks were explored. Hence, we fix  the hyper-parameters shown in Table~\ref{tab:hyper-parameters} to the suggested values in our experiments, which means that only models with 2-stacked LSTM of size 250 are evaluated.

In this study, we also consider the choice of tokenizer and word embedding source as hyper-parameters. To compare the performance of different tokenizers, we tokenize the same split of datasets with different tokenizers and evaluate the overall $F_1$ score over development set. After the best tokenizer for pre-processing patent corpus is determined, we use datasets tokenized by the best tokenizer to train models with different pre-trained word embeddings. The best set of  pre-trained word embeddings for patent corpus is  determined based on the overall $F_1$ score over development set. We then take the best performing tokenizer and pre-trained word embeddings by comparing the marco-average $F_1$ score improvement on both experimental datasets.

\section{Results}

\subsection{Main Results}

\begin{table}[!t]
    \centering
    \scalebox{0.8}{
    \begin{tabular}{l|l|l|l}
        \hline
        \textbf{Tokenizer} & \textbf{BioSemantics} & \textbf{Reaxys} & \textbf{Avg.}\ \\\hline
        OpenNLP   &        89.36 & 89.43 & 89.40\\
        NBIC      &        \textbf{+0.86} & -0.13  & +0.37\\
        ChemTok   &        +0.04 & +1.68 & +0.86\\
        OSCAR4    &        +0.08 & \textbf{+1.86}  & \textbf{+0.97}\\\hline
    \end{tabular}
    }\\
    \caption{Best $F_1$ of EBC-CRF model with different tokenizations on development sets of BioSemantics patent (harmonized set) and Reaxys Gold  with ChemPatent embeddings in use. Recall that the harmonized set of 47 patents is a subset of  BioSemantics, which were annotated by multiple groups (i.e. better annotation quality than remaining patents).}
    \label{tab:results_tokenizers}
\end{table}

\paragraph{Effects of different tokenizers:}\ Table \ref{tab:results_tokenizers} shows that all 3 tokenizers optimized for the chemical domain outperform the baseline general-purpose tokenizer (i.e.\ OpenNLP). The best performance on  BioSemantics  and Reaxys Gold  are achieved by using the NBIC tokenizer (+1.86 $F_1$ score) and the OSCAR4 tokenizer (+0.86 $F_1$ score),  respectively. The best overall tokenizer is OSCAR4 which obtains about 1.0 absolute macro-averaged $F_1$ improvement in comparison to the baseline. 

\begin{table}[!t]
    \centering
    \scalebox{0.8}{
    \begin{tabular}{l|l|l|l}
        \hline
        \textbf{Embeddings} & \textbf{BioSemantics} & \textbf{Reaxys} & \textbf{Avg.}\ \\\hline
        Baseline        &        88.54 &       90.05 &  89.30\\
        PubMed-PMC      &        +0.61 &       +1.03 &  +0.82\\
        Wiki-PubMed-PMC &        +1.24 &  +0.95 & +1.10\\
        ChemPatent      &        \textbf{+1.68} &       \textbf{+1.24} & \textbf{+1.46}\\\hline
    \end{tabular}
    } \\
    \caption{Best $F_1$ of EBC-CRF model with different word embeddings on development sets of BioSemantics patent (harmonized set) and Reaxys Gold  (tokenized by NBIC and OSCAR 4 tokenizer respectively)}
    \label{tab:results_embeddings}
\end{table}

\paragraph{Effects of different sets of word embeddings:}\ 
Table \ref{tab:results_embeddings} shows results obtained by training 
EBC-CRF with different sets of pre-trained word embeddings. On both BioSemantics and Reaxys Gold, it is not surprising that our ChemPatent word embeddings help produce the best performance on the development set, obtaining (on average) a higher $F_1$ score of 1.5 as compared to the baseline embeddings. Specifically, ChemPatent does better than the second best Wiki-PubMed-PMC with about 0.4 improvement. In the rest of the Results section, obtained results are reported with the use of  the OSCAR4 tokenizer and the ChemPatent embeddings  on both experimental datasets.\footnote{OSCAR4 helped produce the highest ``macro-averaged'' improvement on both datasets.} 


\begin{table}[!t]
\centering
\scalebox{.8}{
\begin{tabular}{l|lll}
\hline
\bf Model  &   P & R & $F_1$\\
\hline
tmChem & 72.56 & 78.37 & 75.35 \\
CRFSuite & \textbf{81.93} & 78.38 & 80.12 \\
BiLSTM-CRF + LSTM-char & 79.72 & \textbf{84.42} & \textbf{82.01} \\
\hline
BiLSTM-CNN-CRF & 83.76 &  85.01 &  84.38 \\
EBC-CRF & \textbf{84.30} &  \textbf{87.11} &  \textbf{85.68} \\
\hline
\end{tabular}
}
    \caption{NER scores  on full BioSemantics test set  \citep{akhondi2014annotated}. Results in the first 3 rows were reported in \citet{Maryam2017deep}.  BiLSTM-CRF + LSTM-char denotes the BiLSTM-CRF model with additional LSTM-based character-level word embeddings \citep{Lample:2016}. Recall that our models use the OSCAR4 tokenizer and pre-trained ChemPatent word embeddings.}
    \label{tab:result_comparison}
\end{table}

\begin{table*}[!t]
\centering
\scalebox{0.8}{
\begin{tabular}{l|rr|rrr|rrr|c}
\hline
\multirow{2}{*}{\bf Entity label} & \multicolumn{2}{c|}{\bf Count$^\dagger$} & \multicolumn{3}{c|}{\bf BiLSTM-CNN-CRF} & \multicolumn{3}{c|}{\bf +ELMo} &  \multirow{2}{*}{\bf $\Delta_{F_1}$}\\
\cline{2-9}
                     &  \#   &      \%  &     P  &      R &  $F_1$ &     P &      R &  $F_1$ &   \\
                     \hline
B (Abbreviation)     & 6,558 & 5.78 &  85.90 &  87.02 &  86.46 & 85.78 &  89.98 &  87.83 &   +1.37\\       
C (CAS Number)       & 13    & 0.01 &  54.55 &  46.15 &  50.00 & 57.14 &  61.54 &  59.26 &   +9.26\\       
D (Trademark)        & 2,290 & 2.01 &  62.58 &  61.79 &  62.18 & 66.44 &  71.40 &  68.83 &   +6.65\\       
F (Formula)          & 7,935 & 6.99 &  86.05 &  86.81 &  86.42 & 83.07 &  90.91 &  86.82 &   +0.40\\       
G (Generic)          & 51,313& 45.20&  81.45 &  84.56 &  82.98 & 83.84 &  84.44 &  84.14 &   +1.16\\       
M (IUPAC)            & 39,896& 35.14&  88.40 &  87.77 &  88.09 & 87.25 &  91.20 &  89.18 &   +1.09\\       
MOA (Mode of Action) & 1,137 & 1.00 &  68.97 &  63.32 &  66.02 & 67.62 &  72.74 &  70.08 &   +4.06\\       
R (Registry \#)      & 96    & 0.08 &  55.68 &  51.04 &  53.26 & 65.82 &  54.17 &  59.43 &   +6.17\\       
T (Target)           & 4,290 & 3.78 &  77.77 &  77.32 &  77.55 & 77.21 &  82.68 &  79.85 &   +2.30\\
\hline 
\textbf{Micro Avg.\ }& 113,528 & 100.0&   83.76 &  85.01 &  84.38 & 84.30 &  87.11 &  85.68 & +1.30\\\hline 
\end{tabular}
} \\
\vspace{1em}
(a) BioSemantics\\
\vspace{2em}

\scalebox{0.8}{
\begin{tabular}{l|rr|rrr|rrr|l}
\hline
\multirow{2}{*}{\bf Entity label} & \multicolumn{2}{c|}{\bf Count$^\dagger$} & \multicolumn{3}{c|}{\bf BiLSTM-CNN-CRF} & \multicolumn{3}{c|}{\bf +ELMo} & \multirow{2}{*}{\bf $\Delta_{F_1}$} \\
\cline{2-9}
                                            &       \# &       \%  &     P &      R &  $F_1$ &      P &      R &  $F_1$ & \\\hline         
1 (chemClass)                                &    1,476 & 12.36 & 78.35 &  66.46 &  71.92 &  81.96 &  75.75 &  78.73 &  +6.81\\               
2 (chemClass\textsubscript{biomolecule})     &      951 & 7.96  & 71.86 &  70.50 &  71.17 &  76.27 &  78.76 &  77.50 &  +6.33\\               
3 (chemClass\textsubscript{markush})         &       38 & 0.32  & 42.86 &  47.37 &  45.00 &  42.86 &  47.37 &  45.00 &  +0.00\\               
4 (chemClass\textsubscript{mixture})         &      387 & 3.24  & 76.49 &  59.69 &  67.05 &  74.18 &  64.60 &  69.06 &  +2.01\\               
5 (chemClass\textsubscript{mixture-part})    &      161 & 1.35  & 71.00 &  44.10 &  54.41 &  78.10 &  50.93 &  61.65 &  +7.24\\               
6 (chemClass\textsubscript{polymer})         &      609 & 5.10  & 81.40 &  72.82 &  76.87 &  89.20 &  84.07 &  86.56 & +13.74\\               
7 (chemCompound)                             &    6,988 & 58.53 & 89.02 &  92.01 &  90.49 &  91.01 &  94.58 &  92.76 &  +2.27\\               
8 (chemCompound\textsubscript{mixture-part}) &      904 & 7.57  & 90.02 &  81.86 &  85.75 &  90.63 &  85.62 &  88.05 &  +2.27\\               
9 (chemCompound\textsubscript{prophetics})   &      426 & 3.57  & 18.52 &   2.35 &   4.17 &  77.75 &  79.58 &  78.65 & +74.48\\\hline         
\textbf{Micro Avg.\ }                       &   11,940 &       100.0    & 85.12 &  80.36 &  82.67 &  87.41 &  87.53 &  87.47 &  +4.80\\\hline             
\end{tabular}
}\\
\vspace{1em}
(b) Reaxys Gold\\
    \caption{$F_1$ score with respect to each entity label. ``Count$^\dagger$'' denotes gold-entity counts in test sets.``+ELMo'' denotes scores obtained by EBC-CRF.}
    \label{tab:results_main}
\end{table*}

\paragraph{Final results:}
Table~\ref{tab:result_comparison} compared results reported in \citet{Maryam2017deep} and our approach on the full BioSemantics test set. It is clear that all neural models outperform conventional CRF-based models tmChem and CRFSuite. Our EBC-CRF model outperforms  the BiLSTM-CRF + LSTM-char model with a 3.7 $F_1$ score improvement. Compared to the baseline model BiLSTM-CNN-CRF, the ELMo-based contextualized word embeddings help to produce an $F_1$ improvement of 1.3 points. 

Table~\ref{tab:results_main} details our $F_1$ scores for BiLSTM-CNN-CRF and EBC-CRF with respect to each entity label on both the BioSemantics patent corpus and the Reaxys Gold set. The overall results show that ELMo-based contextualized word embeddings help improve the baseline by 1.3 and 4.8 absolute $F_1$ score on BioSemantics and Reaxys,  respectively.

In BioSemantics patent corpus, we obtain  1+ $F_1$ score improvements on frequent entity labels (i.e.\ $> 3,000$ instances) except for the entity label \textit{Formula}, which has 0.4 absolute improvement. Higher  improvements can be observed on rare entity labels (e.g.\ 4 points on \textit{Mode of Actions}, 6 points on \textit{Registry numbers} and \textit{Trademarks}). The highest improvement at 9 points is found for the most rare entity label \textit{CAS Number}. 

In the Reaxys Gold set, with ELMo  we obtain 2+ $F_1$ score improvements on entity labels \textit{chemCompound}, \textit{chemCompound-mixture part} and \textit{chemClass-mixture}. Higher improvements ($>6$ points) can be seen on some rare entity labels such as \textit{chemClass}, \textit{chemClass-biomolecule}, \textit{chemclass-mixture-part} and \textit{chemClass-polymer}. The  improvements on entity label \textit{chemClass-Markush} and \textit{chemCompound-prophetics} are irregular compared to others. In particular, an absolute $F_1$ improvement of 74+ is achieved on entity label \textit{chemCompound-prophetics}, while we do not find any improvement on \textit{chemClass-Markush}.

\begin{figure*}[ht!]
    \centering
    \begin{tabular}{cc}
        \includegraphics[width=.47500\linewidth]{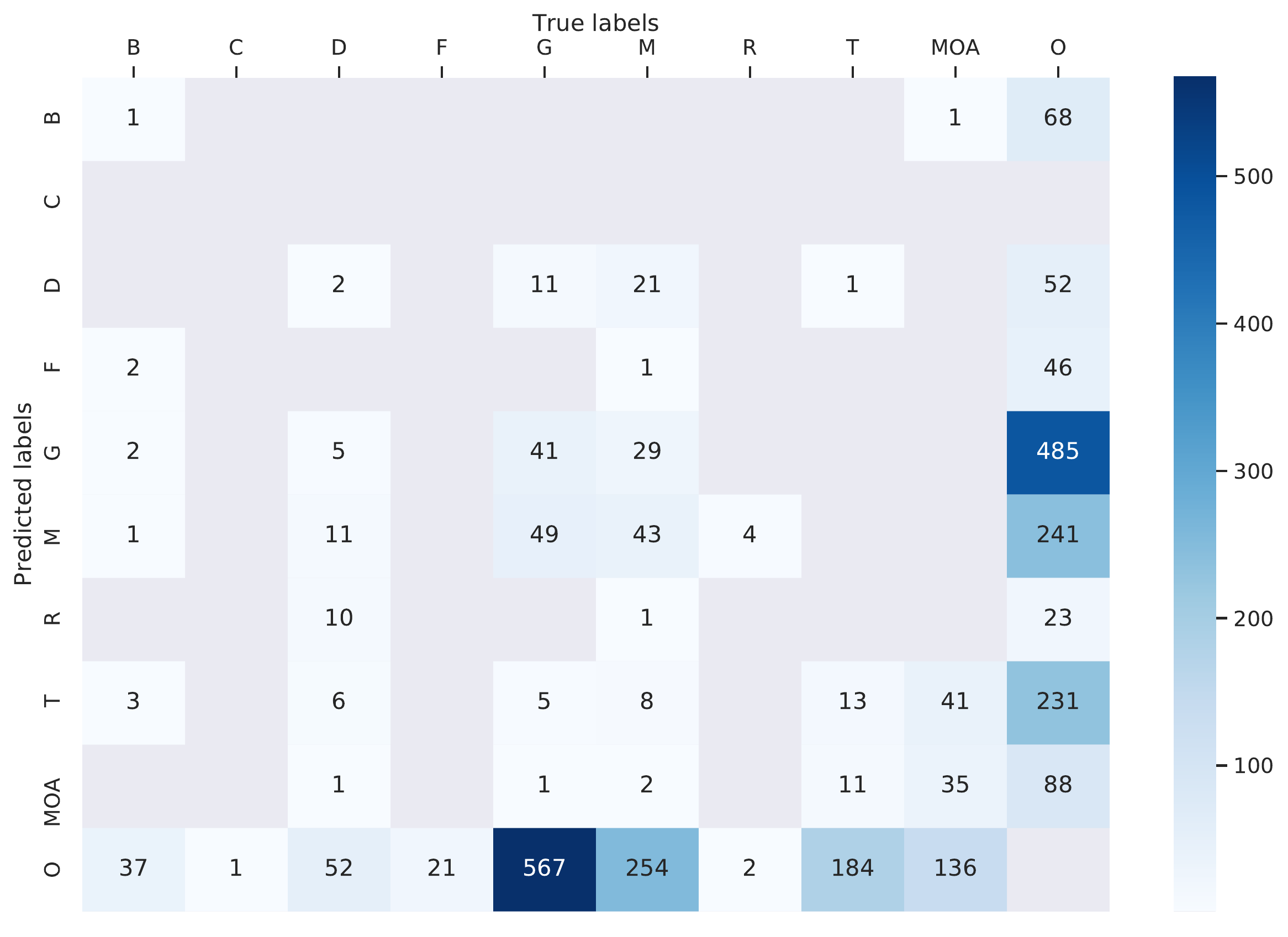}& 
        \includegraphics[width=.47500\linewidth]{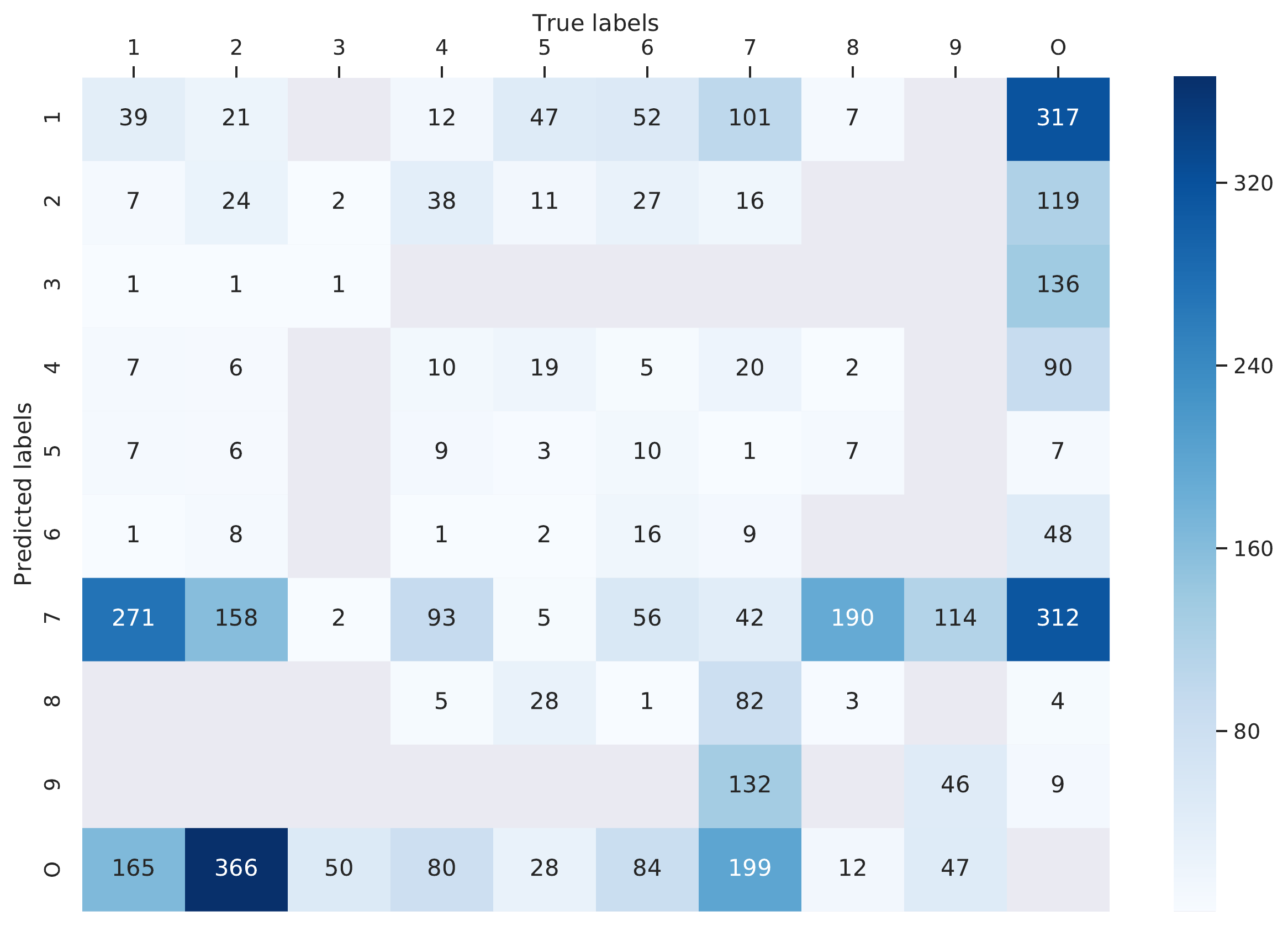} \\
        (a) BioSemantics harmonized set & (b) Reaxys Gold set\\
    \end{tabular}
    \caption{Confusion matrix of
    EBC-CRF 
    model on BioSemantics (harmonized) and Reaxys Gold. $x$-axis: true labels; $y$-axis: predicted labels; numbers on cell where $x=y$ represent confusion between B (Begin) and I (Inside)  tags. In (b) Labels 1--9 are detailed in Table \ref{tab:results_main} (b).
    }
    \label{fig:confusion}
\end{figure*}


\begin{figure*}[ht!]
    \centering
    \includegraphics[width=.99\linewidth]{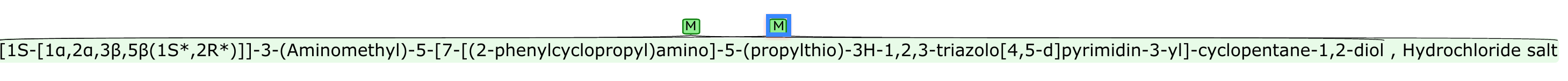} \\
    \vspace{.5em}
    (a) \textit{Salt} describing chemical compound was not detected.
    \includegraphics[width=0.75\linewidth]{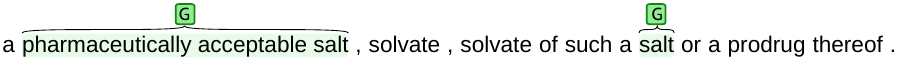} \\
    \vspace{.5em}
    (b) \textit{Salt} describing chemical class being predicted as chemical compound.
    \caption{Example of confusions caused by generic chemical names. (\includegraphics[width=.8em]{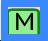}: false negatives, \includegraphics{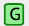}, \includegraphics{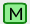}: false positives)}
    \label{fig:example_generic}
\end{figure*}

\subsection{Error Analysis}

To perform error analysis on BioSemantics, we use its harmonized subset. Figure~\ref{fig:confusion}\ (a) shows that  most of the errors are confusions between non-chemical words and generic chemical names (e.g.\ \textit{water, salt, acid}). For example, as illustrated in Figure~\ref{fig:example_generic}\ (a), the word ``\textit{salt}'' which appears at the end of a systematic name should be identified as a part of the systematic name. However, the same word is also widely used to describe a class of chemicals, e.g.\ ``\textit{pharmaceutically acceptable salt}'' in Figure~\ref{fig:example_generic}\ (b).  Disambiguation between  chemical class and chemical compound is a challenging task  even for human annotators, and is thus particularly difficult for a statistical model to learn. The confusion matrix of Reaxys Gold set in Figure~\ref{fig:confusion}\ (b) also supports this point since most confusions are between non-chemical words, chemical classes and chemical compounds.


The Reaxys Gold set has a more complex tag set than the BioSemantics patent corpus, as it assigns separate fine-grained tags for subcategories of chemical classes (\textit{chemClass}) and chemical compounds (\textit{chemCompound}). As illustrated in Table~\ref{tab:results_main}, there is not  sufficient training data for fine-grained sub-category labels. It is difficult for a high complexity neural model to learn characteristics of these sub-category labels and the key difference between the main categories and their subcategories. 
Figure~\ref{fig:confusion}\ (b) shows that 50\% the errors for ``\textit{chemical compound prophetics}'' and 80\% errors for ``\textit{chemical compound mixture part}'' are due to confusion with their parent category ``\textit{chemical compound}''.  

Another typical error observed frequently in BioSemantics and Reaxys is caused by participles.
The most common example is word \textit{'substituted'}.  In ``\textit{substituted or un-substituted alkyl}'', the token ``\textit{substituted}''  refers to a specific chemical compound ``\textit{substituted alkyl}''. Whereas in ``\textit{2-pyridinyl is optionally substituted with 1-3 substituents}'', the token ``\textit{substituted}''  refers to the substitution reaction.

\begin{figure*}
    \centering
    \includegraphics[width=.99\linewidth]{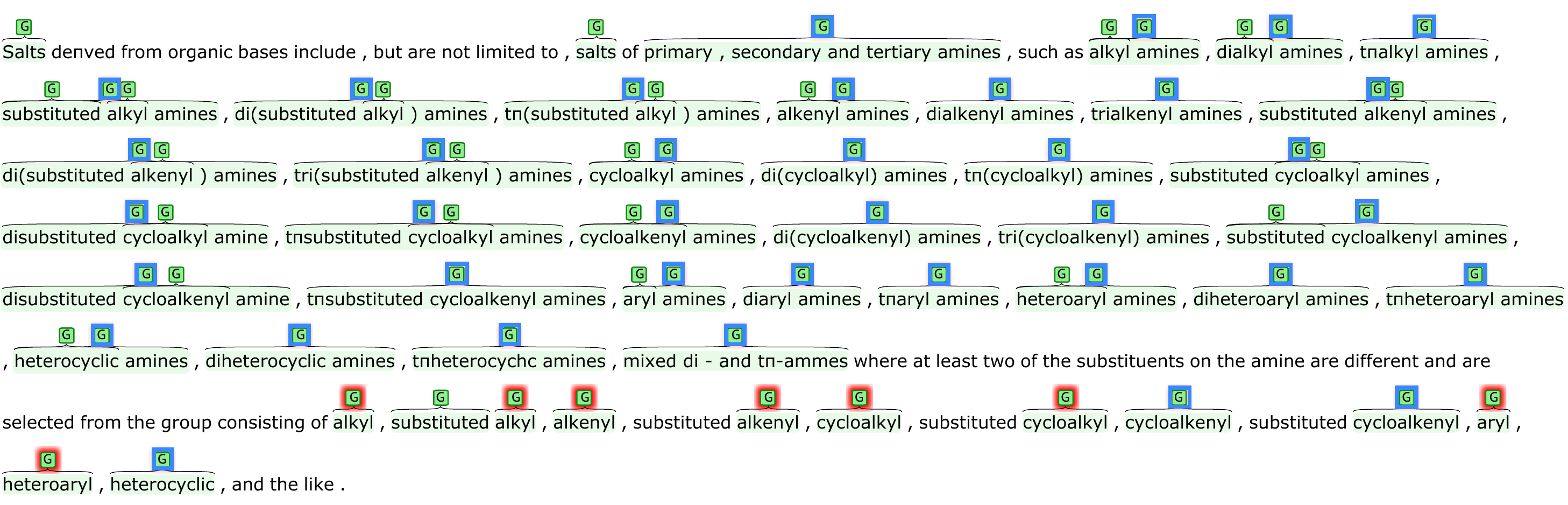}
    \caption{Example of ``chain reaction'' like errors. (\includegraphics[width=.8em]{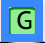}: false negatives, \includegraphics{figs/green-G.pdf}: false positives, \includegraphics{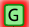}: true positives)}
    \label{fig:example_chain}
\end{figure*}

We also observe that in both patent corpora, there are long sequences of systematic chemical names connected by comma only. Since there are no narrative words between the chemical names in such sequences, it is unlikely that the model can capture any contextual information when tagging them. This can potentially cause a ``chain reaction'' as shown in Figure~\ref{fig:example_chain}, in which all chemical names fail to be recognized when the first chemical name is not tagged correctly.

\subsection{Discussion}

The results in Table~\ref{tab:results_tokenizers} show that all chemical tokenizers outperform the OpenNLP general domain tokenizer. This is not surprising because tokenizers optimized for the chemical domain usually use either rule-based method or gazetteer-based methods to ensure that long systematic chemical names will be treated as a single token instead of being split into several tokens by symbols. 
This is reasonable as the character-level word representation will not be able to capture the morphological structures in a long chemical name if it is split into several tokens.

In the BioSemantics patent corpus, 80\% of all entities are annotated as \textit{Generic} or \textit{IUPAC}. When adding ELMo-based word representations, we obtain smaller improvements in $F_1$ score for \textit{Generic} and \textit{IUPAC} than for remaining entity labels/types. This makes sense, as there are already enough training instances for these two labels in the dataset. By contrast, for rare entity labels with frequencies of less than 2 (e.g.\ \textit{CAS Numbers, Trademarks, Mode of Actions, Registry numbers}), we obtain improvements of 4+ points when exploiting external information conveyed via ELMo.


The global $F_1$ score improvements on both experimental datasets confirm further this observation, viz., that score improvements due to ELMo decrease in inverse proportion to label frequency and training set size. Since the BioSemantics patent corpus contains 10 times more training instances than the Reaxys Gold set, we obtain an absolute improvement of 4.8 on Reaxys Gold set but of 1.3 points on the BioSemantics patent corpus.

Adding ELMo substantially improves the $F_1$ score on \textit{chemCompound-prophetics}. This is because  \textit{chemCompound-prophetics} named entities are all long systematic chemical names which are arranged in lists. Since we replace all tokens longer than 25 characters with ``Long\_Token'' when training ELMo, almost all sentences containing \textit{chemCompound-prophetics} entities appear in the ``Long\_Token'' style. This makes the ELMo-based representations of such long entities almost identical, and particularly easy to predict, thus resulting in an $F_1$ score improvement of 74 points  for \textit{chemCompound-prophetics}. We also observe no improvement for the \textit{chemClass-Markush} label. The Markush structures are figures describing the structure of chemical compounds in which only a few parts/functional groups are labeled. When transforming to text, only the textual labels in the Markush structure are preserved. Thus, it is difficult for ELMo to learn any useful information from the broken Markush structures.

\section{Conclusions}
In this paper, we have made the following contributions towards improved chemical named entity recognition in chemical patents:

\begin{enumerate}[noitemsep]

    \item We improve on the current state-of-art for chemical NER in patents by +2.67 $F_1$ score.

    \item We confirm that tokenizers optimized for chemical domain have a positive effect on NER performance by preserving informative morphological structures in systematic chemical names.
    
    \item We demonstrate that word embeddings pre-trained on an in-domain chemical patent corpus help produce better performance than the word embeddings pre-trained on biomedical literature corpora.
    
    \item We show that chemical NER performance can be improved by using contextualized word representations.
    
    \item We release our ChemPatent word embeddings and an ELMo model trained from scratch on a newly collected corpus of 84K unannotated chemical patents, which can be utilized for downstream NLP tasks on chemical patents.\footnote{\url{https://github.com/zenanz/ChemPatentEmbeddings}}
    
\end{enumerate}

Inspired by the patterns uncovered by our error analysis, our future work on chemical NER will focus on developing models which can be used to support disambiguation of general chemical words. In addition, it would be interesting to explore contextualized word embeddings learned by other neural models such as BERT \citep{devlin2018bert} or OpenAI GPT models \citep{radford2019language} in future work. 


 \section*{Acknowledgments}
 This work was supported by an Australian Research Council Linkage Project grant (LP160101469) and Elsevier BV. We appreciate the contributions of the Content and Innovation team at Elsevier, including Georgios Tsatsaronis, Mark Sheehan, Marius Doornenbal, Michael Maier, and Ralph H\"{o}ssel. 
\bibliography{acl2019}
\bibliographystyle{acl_natbib}





\end{document}